\documentclass[12pt]{article}
\usepackage[a4paper, total={6.7in,10in}]{geometry}
\usepackage{amsmath}
\usepackage{amssymb}
\usepackage{tabularx}	
\usepackage{url}
\usepackage{datetime}

\newdateformat{monthyeardate}{%
  \monthname[\THEMONTH], \THEYEAR}

\def\*#1{\mathbf{#1}}

\newcolumntype{Y}{>{\centering\arraybackslash}X}

\title{\bf A Survey on Neural Network-Based Summarization Methods}
 \author{Yue Dong}
\monthyeardate
 
\begin{document}
\maketitle
\section{Introduction}
Every day, enormous amounts of text are published online and quick access to the major points of these documents is critical for decision making. However, manually producing summaries for such large amounts of documents in a timely manner is no longer feasible. Automatic text summarization, the automated process of shortening a text while reserving the main ideas of the document(s), has consequently became popular. 

Up until recently, text summarization was dominated by unsupervised information retrieval models. In 2014, \cite{ext1_kaageback2014extractive} demonstrated that the neural-based continuous vector models are promising for text summarization.  This marked the beginning of the widespread use of neural network-based text summarization models, because of their superior performance compared to the traditional techniques. 

The aim of this literature review is to survey the recent work on neural-based models in automatic text summarization. This survey starts with the general background on document summarization (Section \ref{sec:background}), including the factors by which summarization tasks may be classified, evaluation issues, and a brief history of the traditional summarization techniques. Section \ref{sec:models} examines in detail ten neural-based summarizers. Section \ref{sec:discussion} discusses the related techniques and presents promising paths for future research, and Section \ref{sec:conclusion} concludes the paper.

\section{Background}\label{sec:background}

\subsection{Summarization Factors}

According to \cite{jones1999automatic}, text summarization tasks can be defined and classified by the following factors: input, purpose and output. 
\subsubsection{Input Factors}
\textbf{Single-document vs. multi-document:} \cite{jones1999automatic} defines this factor as the unit input parameter,  which simply is the number of input
documents that the summarization system takes. 

\textbf{Monolingual, multilingual vs. cross-lingual:} The monolingual summarizers produce summaries that are in the same languages as the inputs, while the multilingual systems can handle the input-output pairs in the same language across several different languages. On the contrary, the cross-lingual summarization systems operate on input-output pairs that are not necessarily in the same language. 

\subsubsection{Purpose Factors}
\textbf{Informative vs. indicative:}

An indicative summary serves as a road-map to convey the relevant contents of the original documents, so the readers can select documents that align with their interests to read further. An indicative summary itself is not supposed to be a substitute for the source documents. On the other hand, the purpose of an informative summary is to replace the original documents as far as the important contents is concerned. 

\textbf{Generic vs. user-oriented:} This factor concerns the coverage of the original documents conditioned on  the potential readers of the summary. 
Generic systems create summaries which consider all the
information found in the documents. In contrast, user-oriented systems produce personalized  summaries that focus on certain information from the source document(s) that are consistent with a user query. 

\textbf{General purpose vs. domain-specific:}
General-purpose summarizers can be used across any  domain(s) with little or no modification.  
On the other hand, domain-specific systems are designed for  processing documents in a specific domain.

\subsubsection{Output Factors}
\textbf{Extractive vs. abstractive:} 
In relation to the source document(s), a summary can either be extractive or abstractive. There is no clear agreement on the definition of the two. 
In this literature review, the definition of \cite{abs4_SeeLM17} is adopted where an extractive summarizer explicitly selects text snippets (words, phrases, sentences, etc.) from the source document(s), while an abstractive summarizer generates novel text snippets to convey the most salient concepts prevalent in the source document(s).

\subsection{Evaluation of Summarization Systems}
Evaluation is critical for developing summarization systems. However, what evaluation criteria should be used for assessing summarization systems still remains unclear due to the subjective aspect of what makes for a good summary. In general, existing evaluation techniques can
be split into either intrinsic or extrinsic
\cite{jones1999automatic}. Intrinsic methods directly evaluate the outcome of a summarization system and extrinsic methods evaluate summaries based on the performance of the down-stream tasks that the system summaries are used for. 

The most prevalent intrinsic evaluation is to compare system-generated summaries (system summaries) with human-created ``gold'' summaries (reference summaries). This allows the use of quantitative measures such as the precisions and recalls in ROUGE \cite{eva1_lin:2004:ACLsummarization}. However, the problem with ROUGE is that people usually disagree on what a ``gold'' summary should be. 

Evaluation methods such as Pyramid \cite{eva2_conf/naacl/NenkovaP04} address this problem and assume that no single best gold summary exists. However, Pyramid is very expensive in terms of the human involvement.

Up to this day,  no single best summarization evaluation method exists and researchers usually adopt the cheap automated evaluation metric ROUGE coupled with human ratings.

\subsubsection{ROUGE \cite{eva1_lin:2004:ACLsummarization}}
Recall-Oriented Understudy for
Gisting Evaluation (ROUGE) are a set of evaluation methods that automatically determine the quality of a system summary by comparing it to human-created summaries. ROUGE-N, ROUGE-L and ROUGE-SU are commonly used in the summarization literatures.

ROUGE-N computes the percentage of n-gram overlapping of system and reference summaries. It requires the consecutive matches of words in n-grams ($n$ needs to be defined and fixed) that is often not the best assumption. 

ROUGE-L computes the sum of the longest in sequence matches of each reference sentence to the system summary. It considers the sentence-level word orders and automatically identify the
longest in-sequence word overlapping without a pre-defined $n$.

ROUGE-SU measures the percentage of skip-bigrams and unigrams overlapping. Skip-bigram consists two words from the sentence with arbitrary gaps in their sentence
order. Applying skip-bigrams without any constraint on the distance between the words usually produce spurious bigram matchings \cite{eva1_lin:2004:ACLsummarization}. Therefore, ROUGE-SU is usually used with a limited maximum skip distance, such as ROUGE-SU4 with maximum skip distance of 4. 

\subsubsection{Pyramid \cite{eva2_conf/naacl/NenkovaP04}}
Instead of matching the exact phrase units as in \cite{eva1_lin:2004:ACLsummarization}, Pyramid tries to score summaries based on semantic matchings of the content units. It works under the assumption that there's no single best summary and therefore multiple reference summaries are necessary for this system.

Given a document with $n$ human created reference summaries $r_1, \ldots, r_n$, Pyramid score of a random summary $s$ is roughly computed as follows:
\begin{enumerate}
\item Human annotation are first required to identify all Summarization Content Units (SCU) in $r_1, \ldots, r_n$ and in $s$, where SCUs are the smallest content unit for some semantic meaning \cite{eva2_conf/naacl/NenkovaP04}.
\item Each SCU is then associated with a weight by counting how many reference summaries in the cluster contain this SCU.
\item Suppose summary $s$ is annotated with $k$ SCUs with weights $w_1,\ldots, w_k$, the pyramid score of $s$ is computed as 
$\sum_{i=1}^{k}w_i/S_{optimal},$
where $S_{optimal}$ is the sum of the $k$ largest SCUs' weights. 
\end{enumerate}    

\subsection{Summarization Techniques}
\subsubsection{Scope of this review}
In this literature review, we primarily consider neural-based extractive and abstractive summarization techniques with the following factors: single document, English, informative, generic and general purpose. As far as I know, related surveys  either investigate the traditional models \cite{medical_summ_survey,das2007survey,nenkova2011automatic} or give little details for neural-based summarizers \cite{gambhir2017recent_survey}.

\subsubsection{Brief History of Pre-Neural Networks Era}
\paragraph{Extractive models}
Most early works on single-document extractive summarization employ statistical techniques based on the ''Edmundsonian paradigm" \cite{medical_summ_survey}. Such algorithms rank each sentence based on its relation to the other sentences by using pre-defined formulas \footnote{These formulas usually don't contain hyper-parameters, and therefore training is not required.} such as the sum of frequencies of significant words (Luhn algorithm\cite{luhn1958automatic}); the overlapping rate with the document title (PyTeaser\cite{pyteaser:Online}); the correlation with salient concepts/topics (Latent Semantic Analysis\cite{gong2001genericLSA}); and sum of weighted similarities to other sentences (TextRank \cite{mihalcea2004textrank}). 

Later works on text summarization address the problem by creating sentence representations of the documents and utilizing machine learning algorithms. These models manually select the appropriate features, and train supervised models to classify whether to include the sentence in the summary. For example, \cite{wong2008extractive_supervised} extracted surface, content, event and relevance features for the sentence representation, and used Support Vector Machines (SVM) and Na\"ive Bayes models for the classification. In addition, sequential models such as Hidden Markov Chains (HMMs) \cite{conroy2001textHMM} were proposed to improve the results by considering the sentence orders in the documents. 

\paragraph{Abstractive models}
The core of abstractive summarization techniques is to identify the main ideas in the documents and encode them into feature representations. These encoded features are then passed to natural language generation (NLG) systems, such as the one proposed in \cite{Reiter:1997:BAN:974487.974490}, for summary generation. 

Most of the early work on abstractive summarization uses semi-manual process of identifying the main ideas of the document(s). Prior knowledge such as scripts and templates are usually used to produce summaries. Thus, the abstractive summary is produced through slot fillings and simple smoothing techniques such as in \cite{j:fru,radev1998generating}.

\section{Neural-Based Summarization Techniques}\label{sec:models}
Since the bloom of deep learning, neural-based summarizers have attracted considerable attention for automatic summarization. Compared to the traditional models, neural-based models achieve better performance with less human involvement if the training data is abundant.  In this section, five extractive and five abstractive neural-based models are examined in details. 

Most neural-based summarizers use the following pipeline: 1) words are transformed to continuous vectors, called word embeddings, by a look-up table; 2) sentences/documents are encoded as continuous vectors using the word embeddings; 3) sentence/document representations (sometimes also word embeddings) are then fed to a model for selection (extractive summarization) or generation (abstractive summarization). 

Neural networks can be used in any of the above three steps. In step 1, we can use neural networks to obtain pre-learned look-up tables (such as Word2Vec, CW vectors, and GloVe). In step 2,  neural networks, such as convolutional neural networks (CNNs) or recurrent neural networks(RNNs), can be used as encoders for extracting sentence/document features. In step 3, neural network models can be used as regressors for ranking/selection (extraction) or decoders for generation (abstraction).

\paragraph{CNNs and RNNs:} CNNs and RNNs are commonly used in neural-based summarizers. Both CNNs and RNNs serve the same purpose: transform a sequence of word embeddings $\mathbf{x}_1,\ldots, \mathbf{x}_T \in \mathbb{R}^d$ to a vector (sentence representation) $\mathbf{s} \in \mathbb{R}^h$.
\begin{itemize}
\item CNNs achieve this purpose by using $h$ filters and  sliding them over the input sequence. Each filter performs local convolution\footnote{The convolution operation used here is basically element-wise matrix multiplication followed by a summation.} on the sub-sequences of the input to obtain a set of feature maps (scalars), then  a global max-pooling-over-time is performed to obtain a scalar. These scalars from the $h$ filters are then concatenated into the sequence representation vector $\mathbf{s} \in \mathbb{R}^h$.

\item RNNs achieve this purpose by introducing time-dependent neural networks.  At the time step $t$, an RNN computes a hidden state vector $\mathbf{h}_t$, which is obtained by a non-linear transformation with  two inputs -- the previous hidden state $\mathbf{h}_{t-1}$ and the current word input $\mathbf{x}_t$:
$$\mathbf{h}_t = f(\mathbf{h}_{t-1}, \mathbf{x}_t).$$
The most basic RNN is called the Elman RNN:
$$\mathbf{h}_t = \sigma(\mathbf{W}_1\mathbf{h}_{t-1}+\mathbf{W}_2\mathbf{x}_{t}).$$
Two other popular RNNs, which address the problem of long-term dependencies by adding extra parameters, are as follows: 

\begin{minipage}[t]{.5\textwidth}
\centering
Gated Recurrent Unit (GRU)
\begin{equation*}
\begin{bmatrix}\*z_t  \\\*r_t \end{bmatrix}=
\begin{bmatrix}\sigma  \\\sigma  \end{bmatrix}
\Bigg(\begin{bmatrix}\*W_1  \\\*W_2  \end{bmatrix}\*h_{t-1}+
\begin{bmatrix}\*W_3  \\\*W_4 \end{bmatrix}
\*x_t\Bigg)
\end{equation*}
$$ \tilde{\*h}_{t}  =\tanh\big(\*W_5(\*r_{t}\odot \*h_{t-1})+ +\*W_6\*x_{t}\big)$$
$$\*h_{t}  =(1-\*z_{t})\odot \*h_{t-1}+\*z_{t}\odot\tilde{\*h}_{t} $$
\end{minipage}
\begin{minipage}[t]{.5\textwidth}\centering
Long short-term memory (LSTM) 
\begin{equation*}
\small\small
\begin{bmatrix}\*i_t  \\\*f_t \\\*o_t\\\*c'_t \end{bmatrix}=
\begin{bmatrix}\sigma  \\\sigma \\\sigma\\tanh \end{bmatrix}
\Bigg(\begin{bmatrix}\*W_1  \\\*W_2 \\\*W_3\\\*W_4 \end{bmatrix}\*h_{t-1}+
\begin{bmatrix}\*W_5  \\\*W_6 \\\*W_7\\\*W_8\end{bmatrix}
\*x_t\Bigg)
\end{equation*}
$$\*c_t = \*f_t \odot \*c_{t−1} +\*i_t \odot \*c'_t$$
$$\*h_t = \*o_t \odot tanh(\*c_t)$$
\end{minipage}
\\
\\
where $\odot$ denotes element-wise matrix multiplication and $\*W_i$ are matrices with the corresponding dimensions. The last hidden state $\*h^T$ is usually used as the sequence representation $\*s=\*h_T\in \mathbb{R}^h$.
\end{itemize}


\subsection{Extractive Models}
Extractive summarizers, which are selection-based methods, need to solve the following two critical challenges: 1)  how to represent sentences; 2) how to select the most appropriate sentences, taking into account of the coverage and the redundancy.
 
In this section, we review  five extractive neural-based summarizers in chronological order. Each summarization system is presented based on its \textit{sentence representation model} and its \textit{sentence selection model}. At the end of this section, the techniques used in the extractive neural-based models are summarized and the models' performance are compared.



\subsubsection{Continuous Vector Space Models \cite{ext1_kaageback2014extractive}}\label{model:ext_1}
\paragraph{Sentence Representation} \cite{ext1_kaageback2014extractive} proposes to represent sentences as continuous vectors that are obtained by either adding the word embeddings or using an unfolding Recursive Auto-encoder (RAE) on word embeddings. The RAE basically combines two text units into one in a recursive manner,  until only one vector (the sentence representation) left. The RAE is trained  in an unsupervised manner by the backpropagation method with the self-reconstruction errors.  The pre-computed word embeddings from Collobert and Weston's model  (CW vectors)  or Mikolov et al.'s model (W2V vectors) are directly used without fine-tuning. 

\paragraph{Sentence Selection}\cite{ext1_kaageback2014extractive} formulates the task of choosing summary $S$ as an optimization problem that maximizes the linear combination of the diversity of the sentences $\mathfrak{R}$ and the coverage of the input text $\mathfrak{L}$:
\begin{equation}
\label{eq:ext1_1}
\mathfrak{F}(S) = \mathfrak{L}(S) + \lambda \mathfrak{R}(S)
\end{equation}
where $\lambda$ is the tread-off between the converge and the diversity. 

According to \cite{ext1_kaageback2014extractive}, this optimization problem is NP-hard. However, there exists fast scalable approximation algorithms with theoretical guarantees if the objective function is submodular \footnote{A function $\mathfrak{F}$ is called submodular on the set $S$ if $\forall s \in S, A \subset B \subset S/\{s\}$ implies $\mathfrak{F}(A+\{s\})- \mathfrak{F}(A) \geq  \mathfrak{F}(B+\{s\})- \mathfrak{F}(B)$. This condition is also called as the \textit{diminishing return property}.}. The authors choose two submodular functions, which are computed  based on sentence similarities, as the diversity function and the converge function, respectively. The objective function is therefore submodular and  an approximation optimization algorithm described in \cite{ext1_kaageback2014extractive} is used for selecting the sentences.

\subsubsection{ CNNLM  \cite{extractive2_2015Yin} }\label{model:ext_2}
\paragraph{Sentence Representation} \cite{extractive2_2015Yin} uses convolutional neural networks (CNNs), similar to the basic CNN model we described previously, on pre-trained word embeddings to obtain the sentence representation.   

The learnable parameters (including the word embeddings) in the CNN are trained by unsupervised learning. The noise-contrastive estimation (NCE) [Mnih and Teh, 2012] is used as the cost function. With this cost function, the model is basically trained as a language model(LM): it learns to discriminate between true next words and noise words. 

\paragraph{Sentence Selection}
Similar as in \cite{ext1_kaageback2014extractive}, the authors frame the sentence selection as a direct optimization problem with the following objective function:
\begin{equation}\label{eq:ext_2}
Q(S) = \alpha \sum_{i \in S}\mathbf{p}_i^2 - \sum_{i,j \in S}\mathbf{p}_i\mathbf{M}_{i,j}\mathbf{p}_j.
\end{equation}
Here, the matrix $\*M$ is obtained by calculating the pairwise cosine similarities of the learned sentence representations. The prestige vector $\mathbf{p}$ is derived by using the PageRank algorithm on $\*M$.

The goal is to find a summary $S$ (as set of sentences) that maximizes the above objective function. Fortunately, equation (\ref{eq:ext_2}) is also submodular (proof in \cite{extractive2_2015Yin}). Therefore, as stated in \cite{ext1_kaageback2014extractive}, a near-optimal solution exists and is presented in  \cite{extractive2_2015Yin}.

\subsubsection{PriorSum \cite{extractive3_cao2015learning}} \label{model:ext_3}
 \paragraph{Sentence Representation}
PriorSum uses the CNN learned features concatenated with document independent features as the sentence representation. Three document-independent features are used: 1) sentence position; 2) averaged term frequency of words in the sentence based on the document; 3) averaged term frequency of words in the sentence based on the cluster (multi-document summarization).

Similar as in \cite{extractive2_2015Yin}, CNNs with multiple filters are used to capture sentence features. However,  PriorSum employs a deeper and more complicated CNN. The CNN used in PriorSum has multiple-layers with alternating convolution and pooling operations. The filters in the convolution layers have different window sizes and two-stage max-over-time-pooling operations are performed in the pooling layers. The parameters in this CNN is updated by applying the diagonal variant of AdaGrad with mini-batches  as described in \cite{extractive2_2015Yin}.

\paragraph{Sentence Selection}
Unlike the previous two extractive neural-based models, PriorSum is a supervised model that requires the gold standard summaries during  training. PriorSum follows the traditional supervised extractive framework: it first ranks each sentence and then selects the top $k$ ranked non-redundant sentences as the final summary.

The authors frame the sentence ranking process as a regression problem. During training, each sentence in the document is associated with the ROUGE-2 score (stopwords removed) with respect to the gold standard summary.  Then a linear regression model is trained to estimate these ROUGE-2 scores by updating the regression weights.

During testing, non-redundant sentences are selected by a simple greedy algorithm. The greedy selection algorithm first ranks all sentences with more than 8 words in descending order based on the  estimated informative scores. The top $k$ sentences are then selected in order as long as the sentence is not redundant with respect to the current summary.  A sentence is considered non-redundant with respect to a summary if more than 50\% of its words do not appear in the summary.

\subsubsection{NN-SE \cite{ext4_cheng-lapata:2016:P16-1}} \label{model:ext_4}
\paragraph{Sentence Representation}
In \cite{ext4_cheng-lapata:2016:P16-1}, sentence representations are obtained by using a CNN followed by an RNN. The CNN extractor, which is similar to the one in \ref{model:ext_2},  has multiple feature maps with different window sizes.  Once sentence representations $(\*s_1, \ldots, \*s_T)$ are obtained by using the CNN sentence extractor, they are fed into an LSTM encoder. The LSTM's hidden states $(\*h_1,\ldots ,\*h_T)$ are then used as the final sentence representations. Comparing to $(\*s_1, \ldots, \*s_T)$, the authors believe $(\*h_1,\ldots ,\*h_T)$ capture the sentence dependency information and  are therefore better suited as sentence representations. 

\paragraph{Sentence Selection}
Similar to \cite{extractive3_cao2015learning}'s work, NN-SE is a supervised model that first scores the sentences and then selects them based on the estimated scores. Instead of using a simple linear regressor as in \cite{extractive3_cao2015learning}, NN-SE  utilizes an LSTM decoder with a sigmoid layer (equation \ref{eq:ext4}) for scoring sentences. During training, the ground truth labels are given (1 for sentences included in the reference summary and 0 otherwise) and the decoder is trained to label sentences sequentially by zeros and ones. 

Given vectors $(\*s_1, \ldots, \*s_T)$ obtained by the CNN  and the LSTM encoder's hidden states $(\*h_1,\ldots ,\*h_t)$, the decoder's hidden states $(\bar{\*h}_1,\ldots ,\bar{\*h}_t)$ are computed as:
\begin{equation}\label{eq:ext4_1}
\bar{\*h}_t = LSTM(p_{t-1}\*s_{t-1}, \bar{\*h}_{t-1})
\end{equation}
where $p_{t-1}$ is the probability that the decoder believes the previous sentence should be included in the summary. The binary decision of whether to include sentence $t$ are modeled by the following sigmoid layer:
\begin{equation}\label{eq:ext4}
p(y(t)=1|D) = \sigma(MLP(\bar{\*h}_t:\*h_t))
\end{equation}
where MLP is a multi-layer neural network.

\paragraph{Joint Training with a Large-Scale Dataset} NN-SE is a sequence-to-sequence model with a CNN+RNN encoder and a LSTM+sigmoid decoder. The encoder (sentence representation model) and the  decoder (sentence selection model) can be jointly trained by the stochastic gradient descent (SGD) method, with the objective of minimizing the negative log-likelihood (NLL):
\begin{equation}\label{eq:ext4-2}
-\text{log}p(y|D, \theta) = -\sum_{i=1}^{m}y_i\text{log}p(y_i|D,\theta).
\end{equation}
Training a sequence-to-sequence summarizer requires a large-scale dataset with extractive labels, i.e., documents with sentences labeled as summary-worthy or not. The authors created a large scale dataset -- the DailyMail dataset -- with about 200K training examples. Each data instance contains an extractive reference summary that is obtained by labeling sentences based on a set of rules such as sentence positions and n-grams overlapping.

\subsubsection{SummaRuNNer \cite{ext5_summarunner}}
\paragraph{Sentence Representation}
SummaRuNNer employs a two-layer bi-directional RNN for sentences and document representations. The first layer of the RNN is a bi-directional GRU that runs on words level: it takes word embeddings in a sentence as the inputs and produces a set of hidden states. These hidden states are averaged into a vector, which is used as the sentence representation. The second layer of the RNN is also a bi-directional GRU, and it runs on the sentence-level by taking the sentence representations obtained by the first layer as inputs. The hidden states of the second layer are then combined into a vector $\mathbf{d}$ (document representation) through a non-linear transformation.

\paragraph{Sentence Selection}
The authors frame the task of sentence selection as a sequentially sentence labeling problem, which is similar to the settings of  \cite{ext4_cheng-lapata:2016:P16-1}. Different from \cite{ext4_cheng-lapata:2016:P16-1}, instead of using another RNN as the decoder, SummaRuNNer uses the hidden states $(\mathbf{h}_1,··· ,\mathbf{h}_m)$ from the second layer of the encoder RNN directly for the binary decision (modeled by a sigmoid function):
\begin{equation} \label{eq:ext5}
P(y_t=1|\mathbf{h}_t,\mathbf{s}_t,\mathbf{d})= \sigma (\mathbf{w}_c \mathbf{h}_t + \mathbf{h}_t^T\mathbf{W}_1 \mathbf{d} -\mathbf{h}_t^T\mathbf{W}_2 \text{tanh}(\mathbf{s}_t)+\*b)
\end{equation}
where $\*b$ includes the information of the sentence's absolute and relative position, as well as the bias. $\mathbf{s}_j$ can be viewed as a ``soft'' summary representation that is computed as the running
weighted sum of sentence representations until time $t$: 
$ \mathbf{s}_t = \sum_{i=1}^{t-1}\mathbf{h}_iP(y_i = 1 |\mathbf{h}_i,\mathbf{s}_i,\mathbf{d})$.

The sigmoid decision layer (\ref{eq:ext5}) and the two-layer encoder RNN (GRU$_{word}$+GRU$_{sent}$) are jointly trained by SGD with the objective function similar to (\ref{eq:ext4-2}) in \cite{ext4_cheng-lapata:2016:P16-1}.

\subsubsection*{Comparison of the Extractive Models and Their Performance} \label{sec:ext_results}
Table \ref{table:ext_compare} compares and summarizes the five extractive models mentioned previously. Almost all these  models are evaluated on DUC2002 dataset and we therefore compare their performance on DUC2002 dataset in Table \ref{table:ext_perf}.
\begin{table}[h]
\footnotesize
\caption{Comparison of the techniques used in the extractive summarizers}
\label{table:ext_compare}
\begin{tabularx}{\textwidth}{|X|X|X|X|X|}
\hline
models & sentence representation & training of sentence representation & sentence selection & training of sentence selection \\ \hline
Continuous Vector Space models (2014)& adding word embeddings or using RAE & no training for adding; RAE is trained in an unsupervise with REs & direct optimization on submodular objectives & no training \\ \hline
CNNLM (2015) & CNN & unsupervised learning with NCE & direct optimization on submodular objectives & no training \\ \hline
PriorSum (2015) & CNN & unsupervised learning with diagonal variant of AdaGrad & sentence ranking by linear regression & supervised learning with ROUGE2 scores \\ \hline
NN-SE (2016)& CNN+RNN & supervised co-train with decoder & sentence ranking from LSTM+sigmoid & supervised learning with SGD and NLL \\ \hline
SummaRuNNer (2017) & GRU+GRU & supervised co-train with decoder & sentence ranking from sigmoid & supervised learning with SGD and NLL \\ \hline
\multicolumn{5}{|l|}{RAE: recursive auto-encoder, REs: reconstruction errors, NCE: noise-contrastive estimation} \\ \hline
\end{tabularx}
\end{table}

\begin{table}[h]
\footnotesize
\caption{Rouge f-scores of the extractive summarizers on the DUC2002 dataset}
\label{table:ext_perf}
\begin{tabularx}{\textwidth}{|X|X|l|l|l|l|}
\hline
models & Extra data used for training (in addition to DUC2002) & rouge1 & rouge2 & rougeL & rougeSU \\ \hline
Continuous Vector Space models (2014) & pre-trained W2V and CW word embeddings & - & - & - & -\\\hline
CNNLM (2015) & pre-trained W2V word embeddings & 51.0 & 27.0 & - & 29.4 \\\hline
PriorSum (2015) & 1. pre-trained CW word embeddings 2. Gigaword for CNN & 36.63 & 8.97 & -& - \\\hline
NN-SE (2016) & DailyMail & 47.4 & 23.0 & 43.5 & - \\\hline
SummaRuNNer (2017) & 1. pre-trained GloVe word embeddings 2. DailyMail & 46.6 & 23.1 & 43.0 & - \\\hline
\end{tabularx}
\end{table}

\subsection{Abstractive Models}
Abstractive summarizers focus on capturing the meaning representation of the whole document and then generate an abstractive summary based on this meaning representation. Therefore, neural-based abstractive summarizers, which are generation-based methods,  need to make the following two decisions: 1)  how to represent the whole document by an encoder; 2) how to generate the words sequence by a decoder.
 
In this section, we review  five abstractive neural-based summarizers in chronological order. Each summarization system is presented based on its \textit{encoder} and its \textit{decoder}. At the end of this section, the techniques used in the abstractive neural-based models are summarized and the models' performance are compared.

\subsubsection{ABS \cite{abs1_RushCW15}}
\paragraph{Encoder}
\cite{abs1_RushCW15} proposes three encoder structures to capture the meaning representation of a document. The common goal of these encoders is to transform a sequence of word embeddings $\mathbf{w}_1, \ldots, \mathbf{w}_T$ to a vector $\mathbf{d}$, which is used as the meaning representation of the document.

\textit{1. Bag-of-Words Encoder:} The first encoder basically computes the summation of the word embeddings appeared in the sequence: $\mathbf{d}_1 = \frac{1}{T}\sum_{i=1}^{T}\mathbf{x}_i$. The word order is not preserved by this bag-of-words encoder.

\textit{2. Convolutional Encoder}: This encoder utilizes a CNN model with multiple alternating convolution and 2-element-max-pooling layers. In each layer, the convolution operations extract a sequence of feature vectors $(\mathbf{u}_1, \ldots, \mathbf{u}_l)$ and the number of these feature vectors are reduced by a factor of two with the 2-element-max-pooling: $\mathbf{\bar{u}}_i = tanh(max\{\mathbf{u}^l_{2i-1},\mathbf{u}^l_{2i}\})$. After $L$ layers of convolution and max-pooling, a max-pooling-over-time is performed to obtain the document representation $\mathbf{d}_2$.

\textit{3. Attention-Based Encoder}: This encoder produces a document representation at each time step based on the previous $C$ words (context) generated by the decoder. At time step $t$, given the inputs' word embeddings $\mathbf{X} = [\mathbf{x}_1, \ldots, \mathbf{x}_m]$ and the decoder's context $\mathbf{y}^{t-1}_C =concat(\mathbf{y}_{t-C}, \ldots, \mathbf{y}_{t-1})$, the encoder produces a document representation (for time step $t$) as follows:
$$\mathbf{d}_3^{t} = \mathbf{p}^T\mathbf{X} \quad where \quad \mathbf{p} \in \mathbb{R}^{m} \propto exp(\mathbf{X}\mathbf{P}\mathbf{y}^{t-1}_C).$$

\paragraph{Decoder}
\cite{abs1_RushCW15} uses a feed-forward neural network-based language
model (NNLM) for estimating the  probability distribution that generates the word at each time step $t$:
$$p(\mathbf{y}_{t}|\mathbf{y}^{t-1}_C,\mathbf{d}^t) \propto exp(\mathbf{W}_1\mathbf{h^t}+\mathbf{W}_2 \mathbf{d}^t) \quad \text{where} \quad \mathbf{h^t}=tanh(\mathbf{W}_3\mathbf{y}^{t-1}_C).$$
\paragraph{Training} In \cite{abs1_RushCW15}, the encoder and decoder are trained jointly in mini-batches. Suppose $\{(x^{(1)}; y^{(1)}),\ldots, (x^{(J)}; y^{(J)})\}$ are $J$ input-summary pairs, then the loss (negative log-likelihood loss (NLL)) based on the parameters $\theta$ is computed as:
\begin{equation} \label{eq:abs1}
NNL(\theta)= -\sum_{j=1}^{J}\text{log}p(y^{(j)}|x^{(j)};\theta) =-\sum_{j=1}^{J}\sum_{t=1}^{T}\text{log}p(y_t^{(j)}|x^{(j)};\theta).
\end{equation}
The training objective is to  minimize the NLL and it is achieved by using mini-batch stochastic gradient descent.

\subsubsection{RAS-LSTM and RAS-Elman \cite{abs2_ChopraAR16}}
\paragraph{Encoder}
The CNN-based attentive encoder used in \cite{abs2_ChopraAR16} is similar to the attentive encoder proposed by \cite{abs1_RushCW15},  except the weights $\alpha_i$ is computed based on the aggregated vectors obtained by a CNN model. At time step $t$, the attention weights are calculated by the aggregated vectors $(\mathbf{z}_1, \ldots, \mathbf{z}_T)$ and decoder's hidden state $\mathbf{h}_{t}$:
$\alpha_{j,t} = exp(\mathbf{z}_j \cdot \mathbf{h}_{t})/ \sum_{i=1}^{t} exp(\mathbf{z}_i \cdot \mathbf{h}_{t})$.
These attention weights are then combined with the inputs' word embeddings to form the document representation $\mathbf{d}_t$ \footnote{$\mathbf{d}_t$, the document representation at time step $t$,  is also called the encoder's context at time step $t$, which is commonly denoted as $\mathbf{c}_t$ in literature.}: $\mathbf{d}_t =\sum_{j=1}^T \alpha_{j,t-1}\mathbf{x}_j$.

\paragraph{Decoder}
\cite{abs2_ChopraAR16} replaces the NNLM model used in \cite{abs1_RushCW15} to a recurrent neural network. Instead of only using the previously-generated $C$ words for decoding as in NNLM, the RNN decoder's hidden state $\mathbf{h_t}$  can keep the information of all the words generated till time $t$. 

The authors propose two decoder models based on the Elman RNN and the LSTM\footnote{How the Elman RNN and LSTM work to produce the hidden states are explained early in section CNNs and RNNs.}. In addition to the previous generated word $\mathbf{y}_{t-1}$ and the previous hidden state $\mathbf{h}_{t-1}$, the Elman RNN and the LSTM take encoder's context vector $\mathbf{d}_t$ (document representation at $t$) as an additional input. For example, the Elman RNN's hidden state is computed as 
$\*h_t =\sigma(\*W_1\*y_{t-1}+\*W_2\*h_{t-1}+\*W_3\*d_t)$.

Once the decoder's hidden state $\mathbf{h}_t$ is computed, it is combined with the document representation $\mathbf{d}^t$ to decide which word to generate \footnote{The words are generated from a pre-fixed dictionary.} at the time step $t$. The decision is modeled by a softmax function, which gives the probability distribution over all the words in the dictionary:
$$\mathbf{P}_t = softmax(\mathbf{W}_4\mathbf{h}_t+\mathbf{W}_5\mathbf{d}_t)$$


\subsubsection{Hierarchical Attentive RNNs \cite{abs3_NallapatiZSGX16}}
\paragraph{Encoder}
\cite{abs3_NallapatiZSGX16} proposes a feature-rich  hierarchical attentive encoder based on the bidirectional-GRU to represent the document. 

\textit{Feature-rich inputs:} The encoder takes the input vector obtained by concatenating the word embedding with additional linguistic features. The additional linguistic features used in their model are parts-of-speech (POS) tags, named-entity (NER) tags, term-frequency (TF) and inverse document frequency (IDF) of the word. The continuous features (TF and IDF) are first discretized into a fixed number of bins and then encoded into one-hot vectors as other discrete features. All the one-hot vectors are then transformed into continuous vectors by embedding matrices and these continuous vectors are concatenated into a single long vector, which is then fed into the encoder. 

\textit{Hierarchical attention:} The hierarchical encoder has two RNNs with a similar structure as in \cite{ext5_summarunner}: one runs on the word-level and one runs on the sentence-level. The hierarchical attention proposed by the authors basically re-weigh the word attentions by the corresponding sentence-level attention. The document representation $\mathbf{d}_t$ is then obtained by the weighted sum of the feature-rich input vectors.
\paragraph{Decoder}
\cite{abs3_NallapatiZSGX16} uses a RNN decoder based on uni-directional GRU, which works similar to the decoder in \cite{abs2_ChopraAR16}. In addition, the following two mechanisms are used in \cite{abs3_NallapatiZSGX16}'s decoder:
\begin{enumerate}
\item \textit{The large vocabulary ‘trick’ (LVT):}
This trick reduces the computation time in the softmax layer by limiting the number of words the decoder can generate from during training. Basically, it defines a  small dictionary in each mini-batch during training. The dictionary only contains the words that are in the source documents of that batch and the most frequent $k$ words in the global dictionary.

\item \textit{Decoder/pointer switch} Using a pointer network, which directly copy words from the source, can improve the summaries' quality by including the rare-words from the source documents. A pointer network can simply be modeled based on the encoder's attention weights where the word with the largest weight is the word for copying. The decision of whether to copy or generate is controlled by a switch, which is modeled by a  sigmoid function $P(s_i = 1) = \sigma(f(\mathbf{h}_{t}, \mathbf{y}_{t-1}, \mathbf{d}_t))$.
\end{enumerate}

\subsubsection{Pointer-Generator Networks \cite{abs4_SeeLM17}}
\paragraph{Encoder}
The encoder of the Pointer-Generator network is simply a single-layer bidirectional LSTM. It computes the document representation $\mathbf{d}_t$ based on the attention weights and the encoder's hidden states, which is exactly the same as the encoder in \cite{abs2_ChopraAR16}.  

\paragraph{Decoder}
The basic building block of \cite{abs4_SeeLM17}'s decoder is a single-layer uni-directional
LSTM. In addition, a decoder/pointer switch similar to \cite{abs3_NallapatiZSGX16} is used for pointing. 

Moreover, the authors propose a coverage mechanism for penalizing repeated attentions on already attended words. This is achieved by using a coverage vector $\*c^t$, which tracks the attentions that all the words in the dictionary has received till time $t$: $\*c^t = \sum^{t-1}_{t'=0} \*a^{t'}$.
The coverage vector is then used for the attention computation at time step $t+1$, as well as in the objective function (acted as a regularizer):
$$L_t = -\text{log}p(w^*_t) + \lambda \sum_{i} \text{min}(\*a_i^t,\*c_i^t)$$
Here, $w^*_t$ is the true label at the time step $t$ and $\lambda$ is a hyperparameter controlling the degree of the coverage regularizer.

\subsubsection{Neural Intra-attention Model \cite{abs_5paulus2017deep}}

\paragraph{Encoder}
\cite{abs_5paulus2017deep} also uses a bi-directional LSTM encoder for modeling the document representation. The model is similar to the encoder in \cite{abs4_SeeLM17}, except the attention scores are computed by linear transformations and a softmax function\footnote{The attention scores in all other models we reviewed are computed by sigmoid functions followed by a softmax function.}, which is called the intra-attention mechanism by the authors. $\mathbf{d}_t$ is then computed based on these intra-attentions and the encoder's hidden states.

\paragraph{Decoder}
A uni-directional LSTM is used as the decoder in \cite{abs_5paulus2017deep}. In addition, the authors employ an intra-attention mechanism on the decoder to prevent generating repeated phrases: a decoder context vector $\mathbf{c}_t$ is computed based on the intra-attentions of the already generated sequence and then used as an additional input for the softmax layer of generating. 
A generator/pointer switch similar to the ones in \cite{abs3_NallapatiZSGX16} and \cite{abs4_SeeLM17} is also employed in the decoder. 

\paragraph{Hybrid Training Objectives}
In terms of the encoder-decoder model, \cite{abs_5paulus2017deep} and \cite{abs4_SeeLM17} are very similar. However, what novel in  \cite{abs_5paulus2017deep} is how the parameters in their model are updated: they use both stochastic gradient descent method and reinforcement learning method to update model parameters with a hybrid training objectives. 

Stochastic gradient descent method (SGD) is used in abstractive summarization models to minimize the negative log-likelihood of the ground-truth values during the training, as explained in the previous models \cite{abs1_RushCW15,abs2_ChopraAR16,abs3_NallapatiZSGX16,abs4_SeeLM17}. We denote this NLL objective as $L_{ml}$. Using SGD to minimize $L_{ml}$ has two shortcomings: 1) it creates a discrepancy during training and testing since there are no ground truth values during testing; 2) optimizing this objective does not always correlate to a high score on the discrete evaluation metric, such as ROUGE scores. 

Therefore, the authors propose to use another objective based on the reinforcement learning method --REINFORCE -- for training: 
$$L_{rl} = (r(\hat{y}-r(y^s))\sum_{t=1}^{T}\text{log}p(y^s_t|y^s_1, \ldots, y^s_{t-1},x)$$
where $y^s$ is obtained by sampling from the $p(y^s_t|y^s_1, \ldots, y^s_{t-1},x)$ at each decoding time step $t$. $\hat{y}$ acts as the REINFORCE baseline, which is obtained by performing a greedy selection rather than sampling at each decoding time step. $r(y)$ is the reward score for an output sequence $y$, which is usually obtained by an automated evaluation method, such as ROUGE.

The authors noticed that optimizing $L_{rl}$ directly would lead to sequences with high ROUGE scores that are ungrammatical. Therefore, a mixed training objective with hyperparameter $\gamma$ is used for balancing the ROUGE score and  the readability of the generated sequence:
$$L_{mixed} = \gamma L_{rl} + (1-\gamma)L_{ml}.$$

\subsubsection*{Comparison of the Abstractive Models and Their Performance} \label{sec:abs_results}
Table \ref{table:abs_compare} compares and summarizes the above five abstractive models. Two large-scale datasets -- the Gigaword dataset and the CNN/DailyMail dataset -- are commonly used as the abstractive summarization benchmarks. We therefore compare the five abstractive models' performance on these two datasets as in Table \ref{table:abs_perf}.
\begin{table}[h]
\footnotesize
\caption{Comparison of the techniques used in the abstractive summarizers}
\label{table:abs_compare}
\begin{tabularx}{\textwidth}{|X|X|X|X|}
\hline
models & encoder & decoder & training \\ \hline
ABS (2015) & \begin{tabular}[c]{@{}l@{}}1. bag-of-words encoder,  \\ 2. CNN,  \\ 3. attention-based encoder\end{tabular} & NNLM & SGD \\ \hline
RAS-LSTM and RAS-Elman (2016) & CNN + attention & Elman RNN or LSTM & SGD \\ \hline
Hierarchical Attentive RNNs (2016) & feature-rich  bidirectional-GRU + hierarchical attention & GRU + LVT + pointer switch & SGD \\ \hline
Pointer-Generator Networks (2017) & bidirectional LSTM + attention & LSTM + pointer switch + coverage mechanism & SGD \\ \hline
Neural Intra-attention Model (2017) & bidirectional LSTM + intra-attention & LSTM + pointer switch + intra-attention & SGD + REINFORCE \\ \hline
\end{tabularx}
\end{table}
\begin{center}
\begin{table}[h]
\footnotesize
\caption{Rouge f-scores of the abstractive summarizers on the Gigaword(G)/CNN-DailyMail(C) datasets}
\label{table:abs_perf}
\begin{tabularx}{\textwidth}{|l|Y|Y|Y|Y|Y|Y|Y|Y|}
\hline
models & \multicolumn{2}{l|}{rouge1} & \multicolumn{2}{l|}{rouge2} & \multicolumn{2}{l|}{rougeL} & \multicolumn{2}{l|}{rougeSU} \\ \hline
 & G & C & G & C & G & C & G & C \\ \hline
ABS (2015) & 29.78 & - & 11.89 & - & 26.97 & - & - & - \\ \hline
RAS-LSTM and RAS-Elman (2016) & 33.78 & - & 15.97 & - & 31.15 & - & - & - \\ \hline
Hierarchical Attentive RNNs (2016 & 35.30 & 35.46 & 16.64 & 13.30 & 32.62 & 32.65 & - & - \\ \hline
Pointer-Generator Networks (2017) & - & 39.53 & - & 17.28 & - & 36.38 & - & 29.4 \\ \hline
Neural Intra-attention Model (2017) & - & 39.87 & - & 15.82 & - & 36.90 & - & - \\ \hline
\end{tabularx}
\end{table}
\end{center}

\section{Discussions and the Promising Paths for Future Research}\label{sec:discussion}
\subsection{Other Related Tasks and Techniques}
\subsubsection{Reinforcement Learning Methods for Sequence Prediction \cite{rl1_bahdanau+al-2017-actorcritic-iclr}}
\cite{abs_5paulus2017deep} shows a promising path of  applying the reinforcement learning (RL) method in abstractive summarization. \cite{abs_5paulus2017deep} applies REINFORCE, which is an unbiased estimator with large variance, for sequence prediction in summarization. 

In \cite{rl1_bahdanau+al-2017-actorcritic-iclr}, the authors apply the actor-critic algorithm, which is a biased estimator with smaller variable, for  machine translation. In addition to the policy network (an encoder-decoder model), they introduce a critic network that is trained to predict the values of output tokens. This critic network is based on a bidirectional GRU and is trained supervisely with the ground-truth labels.

The key difference in the REINFORCE algorithm and the actor-critic algorithm is what rewards the actor uses to update its parameters. REINFORCE uses the overall reward from the whole sequence and only performs the update after obtaining the whole trajectory. The actor-critic algorithm uses the TD errors \cite{rl1_bahdanau+al-2017-actorcritic-iclr} calculated based on the critic network and can update the actor during the generating process. Compared to the REINFORCE algorithm, the actor-critic method has lower variance and faster convergence rate, which makes it a promising algorithm to be used in summarization. 

\subsubsection{Text Simplification \cite{sim1_journals/tacl/XuCN15,sim2_conf/emnlp/ZhangL17}}

The goal of text simplification is to rewrite complex
documents into simpler ones that are easier to understand. This is usually achieved by three operations:
splitting, deletion and paraphrasing \cite{sim1_journals/tacl/XuCN15}. Text simplification can help improve the performance of many natural language processing (NLP) tasks. For example, text simplification techniques can transform long, complex sentences into ones that are more easily processed by automatic text summarizers. 

One challenge of developing text simplification models is the lack of datasets with parallel complex/simple sentence pairs. \cite{sim1_journals/tacl/XuCN15} created a good quality simplification dataset, called the Newsela dataset, for the tasks of text simplification. From their analyses, we could see that the words distribution are significantly different in complex and simple texts. In addition, the distribution of syntax patterns are also very different. These findings indicate that a text simplification model need to consider both the semantic meaning of words and the syntactic patterns of sentences. 

\cite{sim2_conf/emnlp/ZhangL17} propose a sequence-to-sequence model with attentions based on LSTMs for text simplification. This encoder-decoder model, called  Deep REinforcement
Sentence Simplification (DRESS), is trained with the reinforcement learning method that optimizes a task-specific discrete reward function. This discrete reward function encourages the outputs to be simple, grammatical, and semantically related to the inputs. Experiments on three datasets demonstrate
that their model is promising for text simplification tasks.

\subsection{Discussions}
In summarization, one critical issue is to represent the semantic meanings of the sentences and documents. Neural-based models display superior performance on automatically extracting these feature representations. However, deep neural network models are neither transparent enough nor integrating with the prior knowledge well. More analysis and understanding of the neural-based models are needed for further exploiting these models. 

In addition, the current neural-based models have the following limitations: 1) they are unable to deal with sequences longer than a few thousand words due to the large memory requirement of these models; 2) they are unable to work well on small-scale datasets due to the large amount of parameters these models have; 3) they are very slow to train due to the complexity of the models.

There are many very interesting and promising directions for future research on text summarization.  We proposed two directions in this review: 1) using the reinforcement learning approaches, such as the actor-critic algorithm, to train the neural-based models; 2) exploiting techniques in text simplification to transform documents into simpler ones for summarizers to process.

\section{Conclusion}\label{sec:conclusion}
This survey presented the potential of neural-based techniques in automatic text summarization, based on the
examination of the-state-of-the-art extractive and abstractive summarizers. Neural-based models are promising for text summarization in terms of the performance when large-scale datasets are available for training. However, many challenges with neural-based models still remain unsolved. Future research directions such as adding the reinforcement learning algorithms and text simplification methods to the current neural-based models are provided to the researchers.


\bibliographystyle{apalike}
{\footnotesize
\bibliography{sample}}
\end{document}